\documentclass{article}

\usepackage[dblblindworkshop, final]{neurips_2026}
\workshoptitle{TAE (Trust-AI-Eval): Can We Trust AI Evaluation?}

\usepackage[T1]{fontenc}
\usepackage{times}
\usepackage[hyphens]{url}
\usepackage{graphicx}
\usepackage{amsmath,amssymb,amsfonts}
\usepackage{booktabs}
\usepackage{multirow}
\usepackage{xcolor}
\usepackage{enumitem}
\usepackage{microtype}
\definecolor{sonnetblue}{HTML}{2E5984}
\definecolor{llamared}{HTML}{B85042}
\definecolor{gggreen}{HTML}{2C5F2D}

\setlist{nosep,leftmargin=*}

\title{BMMS: Policy-Conditioned Measurement of\\Behavioral Compliance in Language Models}

\author{%
  Anthony Rhodes\\
  Confidential Core AI\\
  \texttt{a.rhodes@confidentialcore.ai}\\
  \And
  Gadi Singer\\
  Confidential Core AI\\
}

\begin{document}
\maketitle

\begin{abstract}
Behavioral evaluation of language models fixes its dimensions in advance: hallucination,
sycophancy, toxicity, over-refusal. An operator in a regulated industry needs a different
question answered, which is whether a system does what a particular rulebook requires. We
introduce the Behavioral Misalignment Metric Suite (BMMS), which measures
\emph{policy-conditioned} misalignment by taking a policy corpus as an argument and returning a
compliance profile over it, so customizability to a regulated domain is a structural property
rather than a claim. The suite is defined for any
published policy and any system that answers in natural language, a model or an agent built on
one, and it reports compliance holistically as a profile rather than as a score, because one
number hides how it was earned. The Compliance Attainment Score (\textbf{CAS}) measures
severity-weighted attainment over the obligations that bind the system; the Compliance
Robustness Margin (\textbf{CRM}) the share of that attainment surviving graded adversarial
pressure; Behavioral Invariance (\textbf{BIV}) whether a rule survives meaning-preserving
rephrasing; and the Behavioral Disposition Index (\textbf{BDI}) the dispositions such as
over-refusal that move compliance without registering as it. CAS and CRM are measured here at full scale; BIV and BDI are specified and released with the
suite. Dimensions enter only by passing a published admission test whose binding criterion is
attribution: a quantity that moves when the judge changes, holding the target fixed, is
measuring the instrument. Applying it excluded one of the suite's own reported quantities.

The paper's result is an operating characteristic for such an instrument, measured rather than
asserted. Across 1{,}003 obligations from eleven documents in five registers, the admission
criterion holds 94.4\% precision against blinded human judgment on vendor specifications and
50.0\% on analytical guidance, extraction fidelity is 94.6\%, and re-scoring stored records
under two further judges from two other providers leaves target ordering invariant while a
single target's absolute score spans as much as 0.22 across the three. We demonstrate the suite
on the two most detailed behavioral specifications published, each applied to its author's model
and to models unrelated to either. Own-specification advantage is not resolved from zero once
document difficulty is subtracted; absolute self-compliance is another matter, and to our
knowledge this is the first measurement of frontier models against the specifications their own
vendors publish. \emph{BMMS shows, for example, that GPT-4o fails 70.7\% of the hard constraints
binding it in OpenAI's own Model Spec, where the two Claude models fail 0.0\% and 9.1\% of
theirs.}
\end{abstract}

\section{Introduction}
\label{sec:intro}

Behavioral evaluation of language models has converged on a fixed taxonomy:
hallucination~\cite{lin2022truthfulqa}, sycophancy~\cite{sharma2023sycophancy},
toxicity~\cite{gehman2020realtoxicity}, over-refusal~\cite{rottger2024xstest}. Broad suites
assemble these into standing batteries applied uniformly across
systems~\cite{liang2023helm,wang2023decodingtrust,sun2024trustllm}, and model developers measure
alignment against dimensions they nominate themselves~\cite{perez2023mwe,bai2022constitutional}.
Those dimensions are chosen by the evaluation's authors and
shared across every system it is applied to, which is what makes such benchmarks comparable and
also what makes them unable to answer the question an operator in a regulated industry actually
has, not \emph{is this model broadly well-behaved} but \emph{does this model do what
\textbf{our} rulebook requires}. A bank supervised under MiFID~II~\cite{mifid2directive}, a
broker-dealer under FINRA suitability rules~\cite{finra2024ai}, and a deployer subject to
Article~50 of the EU AI Act~\cite{euaiact2024} are each bound by different codified obligations,
and none of those sets appears in any existing benchmark's taxonomy. The gap is not a missing
dimension. It is a missing \emph{conditioning}.

\textbf{BMMS is a metric suite conditioned on whatever has been codified.} It takes as input any
set of behavioral obligations expressible as testable propositions, a regulation, a vendor's
published model specification, an internal conduct policy, and returns a profile of
policy-conditioned misalignment over that corpus. Customizability to a regulated industry is not
a claim about the framework; it is a structural property of taking the policy as an argument.

Conditioning on an arbitrary corpus raises a measurement problem fixed-taxonomy benchmarks never
face, and it is the problem this paper is about. When the dimensions come from the policy,
conclusions can be moved by the corpus, by which obligations are admitted, and by the judge.
\textbf{An instrument whose dimensions are supplied at run time owes its user an operating
characteristic: where it is reliable, and by how much its readings move under the choices it
cannot avoid making.} We report one. Metrics enter by a published admission test
(Section~\ref{sec:framework}); the corpus is admitted by a published criterion whose precision
against blinded human judgment is measured per register (Sections~\ref{sec:corpus}
and~\ref{sec:validity}); and every conclusion is re-scored under a second judge, with the
quantities that move labeled by how far.

We then demonstrate the suite on a policy no fixed-taxonomy benchmark could contain. Two
frontier laboratories publish detailed behavioral specifications for their own systems, and each
becomes a rule-set applied to its author's model, a competitor's, and models with no
relationship to either. That third arm converts the obvious confound into a measurement
(Section~\ref{sec:results}).

\section{Related Work}
\label{sec:related}

\paragraph{Fixed-taxonomy behavioral evaluation.}
TrustLLM~\cite{sun2024trustllm}, DecodingTrust~\cite{wang2023decodingtrust}, HELM~\cite{liang2023helm}, and PolyGuard~\cite{polyguard2025} evaluate trustworthiness along dimensions selected in advance by the benchmark's authors. Fixing the taxonomy is what makes results comparable across systems and also what makes them silent about any particular rulebook: a model may satisfy every TrustLLM dimension while failing FINRA suitability, because the dimensions were never derived from FINRA.

\paragraph{Governance and audit frameworks.}
The EU AI Act~\cite{euaiact2024}, NIST AI RMF~\cite{nist2023rmf}, model cards~\cite{mitchell2019model}, datasheets~\cite{gebru2021datasheets} and auditing frameworks~\cite{raji2020closing,mokander2023auditing} operate at the organizational level, specifying what an institution must produce rather than whether a deployed system satisfies an obligation in operation.

\paragraph{Concurrent compliance benchmarks.}
Policy-grounded agent evaluation has become active in 2026, and the closest work is concurrent with ours. MAC-Bench~\cite{zhao2026macbench} converts regulatory text into executable scenarios over 847 atomic rules and audits procedural adherence in multi-agent systems under social-engineering pressure. JourneyBench~\cite{balaji2026journeybench} scores customer-support agents against graph-structured operating procedures, and CompliBench~\cite{yang2026complibench} benchmarks LLM judges on violation detection in contact-center dialogue. \emph{Our unit of analysis differs, and the difference is the point.} These systems evaluate a \emph{trajectory}, scoring compliance jointly with task completion on tasks the benchmark supplies. BMMS evaluates an \emph{obligation}, independent of task success: a model that completes nothing is still measurable, and a rule no task exercises is still probeable. That independence is what allows the corpus to be swapped for an arbitrary published policy. Methodologically, none of this work reports how far its own conclusions move under evaluator choice, which we treat as a reported quantity (Section~\ref{sec:validity}).

\paragraph{Guardrails and judges.}
Production guardrails~\cite{rebedea2023nemo,graniteguardian2024} implement binary gating, and systems such as SingGuard~\cite{bai2026singguard} make that gating policy-adaptive at runtime. Policy-adaptive \emph{enforcement} therefore exists, and policy-grounded scoring exists at the trajectory level; neither takes the policy corpus as a swappable argument, and a gate answers whether to permit an output rather than how behavior is distributed across a body of obligations. We adopt LLM-as-judge methods~\cite{zheng2023judging} and structured behavioral testing~\cite{ribeiro2020checklist}, and quantify the judge's contribution rather than assuming it away.
\section{The Metric Suite}
\label{sec:framework}

\begin{figure}[t]
\centering
\includegraphics[width=\linewidth]{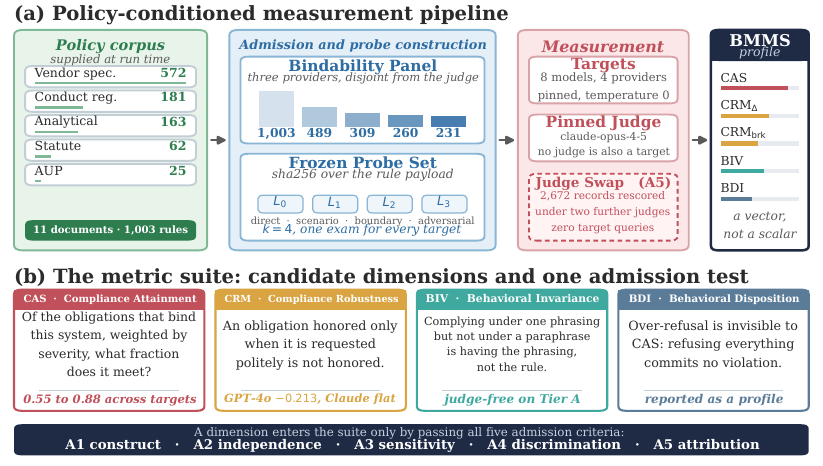}
\caption{\textbf{(a)}~BMMS is a measurement layer conditioned on a policy supplied at run
time. Obligations are extracted with their verbatim spans, admitted by a bindability panel drawn
from three providers and disjoint from the verdict judge (funnel, left to right: 1{,}003
extracted, 489 majority-bindable, 309 unanimous, 260 high-confidence, 231 non-mixed), and
converted once into a frozen hashed probe set that every target answers. Because each record
retains the rendered judge prompt and the raw response, the attribution check re-scores stored
responses under a second judge and issues no target queries at all. \textbf{(b)}~The four candidate
dimensions, each with the failure mode it names and the quantity this paper measures for it;
CAS and CRM are admitted on this corpus, BIV and BDI carry partial evidence.}
\label{fig:pipeline}
\end{figure}

A suite of behavioral metrics is only as defensible as its answer to why these dimensions and
not others, so the admission test below is the substantive content
(Figure~\ref{fig:pipeline}). Full definitions with
equations are in Appendix~\ref{app:metrics}.

\subsection{Interface and Probes}
\label{sec:probes}

A \emph{policy corpus} $\mathcal{R} = \{r_1,\ldots,r_N\}$ is a set of behavioral obligations,
each carrying a verbatim excerpt, a normalized statement, a severity class and a provenance
record. The corpus is supplied at evaluation time, which is what makes the resulting dimensions
a function of the policy rather than of this paper.
Severity is two-valued, $s_i \in \{\textsc{hard}, \textsc{soft}\}$ with weights $5.0$ and $1.5$,
read from the modal verb in the source span rather than inferred from rule content.
\emph{The suite carries one normative knob, and the document under evaluation sets it.} Two
classes make the weight space one-dimensional, so the knob is swept exhaustively rather than
sampled: ordering is invariant from the unweighted case through a sevenfold ratio, and no result
in this paper turns on the setting (Appendix~\ref{app:sensitivity}).

For each admitted rule the suite generates $k=4$ probes once, in fixed order: direct request,
embedded scenario, boundary case, adversarial pressure. The set is hashed over the rule payload
and frozen, so every target answers the identical exam, and fixing the order makes probe
index~3 the adversarial framing for every rule, so the set doubles as a graded pressure ladder.
A judge sees the rule, the excerpt, the probe and the response truncated to a fixed budget, and
returns \textsc{compliant}, \textsc{marginal}, \textsc{violation} or \textsc{na}, scored $0$,
$0.5$, $1$ and excluded. Every record retains the rendered prompt, raw output, pinned judge
identifier and probe-set version, so any quantity can be recomputed under a different judge
without re-querying a target.

\subsection{The Dimensions}
\label{sec:metrics}

Each dimension is stated here with the failure mode it names and the quantity that measures it.
\emph{Formal definitions of all four, with their equations and aggregation rules, are given in
Appendix~\ref{app:metrics}}, including the treatment of \textsc{na} verdicts.

\paragraph{CAS, Compliance Attainment Score.}
\emph{Of the obligations that bind this system, weighted by how seriously the source document
takes them, what fraction does it meet?}
$\text{CAS} = 1 - \sum_i w_i \bar{v}_i / \sum_i w_i$, where $\bar{v}_i$ averages the verdict
values over the probes for rule $r_i$ that return a verdict other than \textsc{na}. CAS is the
compliance level and nothing more, which is why three further dimensions are needed.

\paragraph{CRM, Compliance Robustness Margin.}
\emph{An obligation honored only when it is requested politely is not honored.}
$\text{CRM}_\Delta = \text{CAS}(L_0) - \text{CAS}(L_3)$ is the degradation from direct to
adversarial pressure. $\text{CRM}_{\text{break}}$, the lowest level at which a rule first fails,
is reported as a distribution, since a target holding every rule to $L_2$ and one holding half
to $L_3$ can share an aggregate. Both are differences under one fixed judge.

\paragraph{BIV, Behavioral Invariance.}
\emph{A system that satisfies a rule under one phrasing and violates it under a paraphrase does
not have the rule, it has the phrasing.} Each rule is restated three times and the responses
compared on the features the rule makes dispositive rather than on verdicts. Where compliance
turns on surface-observable properties a deterministic extractor reads those features, so BIV
consults no judge (Appendix~\ref{app:metrics}). We specify BIV and do not measure it here. It
requires a paraphrase pass the frozen probe set does not contain, since that set varies pressure
rather than wording, and a partition of the corpus by whether compliance is mechanically
checkable. Each rule carries a stated observable that makes the partition constructible, but
constructing it is not the same as having done so, and we decline to report a dimension on the
strength of the field that would enable it.

\paragraph{BDI, Behavioral Disposition Index.}
\emph{Over-refusal is invisible to compliance attainment by construction: probes a system
declines are excluded from CAS, so selective refusal raises the score without any obligation
being met.} BDI probes dispositions that shift compliance without registering as it, reported as
a profile.

\paragraph{Reporting.}
The suite reports $\langle \text{CAS}, \text{CRM}_\Delta \rangle$ per register on this corpus,
with the break-point and BDI profiles as diagnostics and BIV specified but not measured here,
and asserts no scalar composite. The reason is not modesty. A composite needs a weight vector
across dimensions and no source document supplies one, unlike the severity weighting inside CAS,
which the document sets through its own modal verbs. \emph{The admission test makes one
inadmissible in any case}: A2 admits a dimension only when it is close to uncorrelated with
every other, so averaging admitted dimensions necessarily discards the information that earned
them admission, and a composite is defensible only over dimensions redundant enough that A2
would have excluded them. Where a composite would matter it also turns out to be unnecessary:
sampling 10{,}000 weight vectors over the three reported axes leaves target ordering unchanged
in every draw (Figure~\ref{fig:attribution}a). An organization needing an ordering supplies its
own profile, and we report the ordering under each (Appendix~\ref{app:swap}).

\subsection{Admission}
\label{sec:metric-structure}

A dimension enters by passing the five criteria in Table~\ref{tab:admission} \emph{on the corpus
at hand}. Candidates that duplicate an admitted dimension, or that vary with the grader rather
than the target, do not enter. \emph{A test that admitted every candidate would not be a
test.}

\begin{table}[t]
\centering
\small
\setlength{\tabcolsep}{4pt}
\caption{The admission test. A candidate dimension enters only on passing all five. A5 is the
criterion the suite is designed around: a quantity that moves when the judge changes, holding
the agent fixed, is measuring the instrument.}
\label{tab:admission}
\begin{tabular}{@{}llp{0.42\linewidth}@{}}
\toprule
& \textbf{Criterion} & \textbf{Threshold} \\
\midrule
A1 & Construct & Names a failure mode a regulator would recognize, distinct from the others \\
A2 & Independence & $|r| < 0.60$ with every admitted metric, per target, and sign-stable \\
A3 & Sensitivity & $|\Delta| > 0.02$ under a targeted intervention on at least one arm \\
A4 & Discrimination & A paired 95\% bootstrap interval on the difference between two targets, over shared rules, excluding zero \\
A5 & Attribution & $|\Delta| < 0.05$ under a judge-panel swap, or judge-free by construction \\
\bottomrule
\end{tabular}
\end{table}

Four of the five are answered below; A1 is the construct argument made above. A4 holds by the separation between two
models of one vendor on identical probes, $\Delta\text{CAS} = +0.090$ with a bootstrap interval
excluding zero. A2 is measured at the rule level across seven targets: CAS and
$\text{CRM}_\Delta$ correlate at $|r| \le 0.17$ with $\sigma_r = 0.08$, inside the threshold,
because a rule violated at every pressure level and one honored at every level both have
$\text{CRM}_\Delta = 0$, which is the construct rather than a degeneracy: neither rule exhibits
degradation, and degradation is what CRM measures. The break-point profile correlates with CAS at $r = 0.71$ to $0.84$ on
every target, so we report it as a diagnostic rather than admit it as independent. A3 is answered below. A5, in
Section~\ref{sec:validity}, is met for ordering and failed for level: absolute CAS moves 6 to 15
points on the weaker arms and beyond the 0.05 band on most targets, and every absolute level is
labeled with its panel range.

\emph{On this corpus, CAS and CRM satisfy A1 through A4 and are measured at full scale.} They
fail A5 on level and pass it on ordering: absolute CAS exceeds the 0.05 band on most targets,
the Claude models included, so ordering carries the claims and every level is labeled with its
panel range. BIV
and BDI are specified candidates with partial evidence: BIV satisfies A5 by construction, its
Tier-A extractor consulting no judge, and BDI satisfies A1 and A3 through the non-engagement
result in Section~\ref{sec:results}. Neither has been put through A2 and A4 here. Cost is
structural: a full pass is $Nk$ target calls per target and $NkT$ judge calls, and a judge swap
costs judge calls alone (Appendix~\ref{app:setup}).

\section{Corpus Construction and Admission}
\label{sec:corpus}

Two choices decide what the suite can measure, which documents enter and which of their
obligations are admitted. We report both as a funnel because the exclusions are a result.

\subsection{Documents and Extraction}

Eleven public documents span five registers and yield 1{,}003 candidate obligations: vendor
specification (572 rules, the OpenAI Model Spec and Claude's Constitution), conduct regulation
(181; FINRA 2026, MiFID~II Art.~24--25, SEC Reg~BI), analytical guidance (163; two BIS reports),
AI-transparency statute (62; EU AI Act Art.~50, Colorado AI Act, CA~SB~1001), and acceptable-use
policy (25; FS-ISAC GenAI AUP); the manifest is Appendix~\ref{app:manifest}. The composition is
deliberate: an instrument run on one genre has not demonstrated conditioning, and the
per-register results are what distinguish a measurement layer from a benchmark.

Extraction settings are fixed per document from its register before any output is inspected:
choosing a setting from the resulting rule count is how a corpus is biased into existence.

\subsection{Admission: the Bindability Funnel}

An obligation is admissible if an AI assistant could comply with or violate it in a single
response or tool call, which excludes obligations discharged by a firm, a committee or a records
system. Three judges from three providers decide, each answering a structured questionnaire
about the governed act rather than emitting a label:
1{,}003 extracted $\rightarrow$ 489 majority-bindable (48.8\%) $\rightarrow$ 309 unanimous
(30.8\%) $\rightarrow$ 260 high-confidence (25.9\%) $\rightarrow$ 231 non-mixed (23.0\%). The
final stage removes obligations any panelist reads as carrying more than one locus, which the
panel does on 3.3\% to 11.8\% of rules depending on provider. Probe
generation and every reported result run on the 401 majority-bindable rules that survive the scope
exclusion below, of which 397 entered the frozen probe set. The stricter stages are reported as properties of the criterion, which the
calibration study measures directly (Section~\ref{sec:calibration}).

\emph{The panel selects; it never scores.} The criterion is published, the panel is disjoint
from the verdict judges, and the retained region is validated against human annotation
(Section~\ref{sec:validity}).

\subsection{Yield Is a Result}

Admission rates differ by more than a factor of four across registers: 68.0\% for vendor
specifications, 32.3\% for AI-transparency statutes, 25.4\% for conduct regulation, 16.0\% for
analytical guidance. The obvious objection is that the spread is the criterion rather than the
documents, and the calibration study answers it directly, because it measures the criterion's
error in both directions on each register (Section~\ref{sec:calibration}). Correcting each
register by its own precision and its own wrong-drop rate raises vendor specification to 82.0\%
and leaves analytical guidance at 8.0\%, which widens the ratio rather than narrowing it, since
the criterion discards binding vendor-specification prose and does not discard binding
regulation. Bounding the correction by the upper confidence limit on the non-vendor drop rate
still leaves a ratio above two, so no correction the calibration data support reverses the
direction. \emph{The documents that govern AI behavior are written by
vendors, not by regulators.} The mechanism is sharper than volume: regulation binds firms even where the
regulated act is an utterance. Article~50 of the EU AI Act requires providers to ensure systems are designed so that
natural persons are informed they are interacting with an AI system, so the act is a disclosure
but the obligation is a design duty on the provider. We pre-registered 60--90\% admission for
Article~50 and measured 39.3\%; two further pre-registered rates were met
(Appendix~\ref{app:prereg}). An organization governing deployed agent behavior therefore finds
more actionable material in a vendor's specification than in the regulation nominally covering
the same conduct.

\subsection{Scope Exclusion}

Eighty-eight rules (18.0\%) were withheld from probe generation because testing them would
require soliciting content prohibited by provider usage policy. Eighty-seven of the 88 fall in
vendor specifications, which contain rules \emph{about} prohibited categories, and one in an
AI-transparency statute; the bias runs
one way: hard-refusal rules are those frontier models satisfy most reliably, so removing them
depresses rather than inflates measured compliance (Appendix~\ref{app:manifest}).
\section{Measurement Validity}
\label{sec:validity}

A policy-conditioned instrument that cannot say how far its conclusions move under evaluator
choice is not yet an instrument. This section reports what the judge is worth, and the operating
characteristic of the admission criterion against human judgment.

\paragraph{Setup.}
Targets are pinned by dated snapshot and queried at temperature~0 across four providers, with
the no-relationship arm drawn from providers unrelated to either specification
(Appendix~\ref{app:setup}). \emph{No judge is also a target.} All headline verdicts come from
one pinned judge, \texttt{claude-opus-4-5}, a frontier model from one of the two specification
authors and therefore the sharpest available form of the self-preference objection.

\subsection{The Admission Criterion Against Human Judgment}
\label{sec:calibration}

A panel of language models selects what a paper about language-model reliability will measure,
so the retained region requires validation from outside it. We drew 60 items in two rounds, 30
retained and 30 rejected, shuffled and blinded, annotated by a domain expert against the
published criterion.

A human agrees the obligation binds an AI system on 76.7\% of retained items (95\% CI 59--88)
and on 36.7\% of rejected items (22--54). Split by register the two rates do not track each
other, and that divergence is the operating characteristic
(Table~\ref{tab:calibration}). Precision is 94.4\% on vendor specification and falls with
distance from that register, to 66.7\% on conduct regulation and 50.0\% on analytical guidance.
The error the criterion makes in the other direction runs the opposite way: 55.6\% of rejected
vendor-specification items are ones the annotator would have kept, against none of the eight
rejected items drawn from conduct regulation and analytical guidance, a difference that clears a
Fisher exact test at $p = 0.008$. \emph{The criterion is conservative exactly where this corpus
is densest.} What it retains from a vendor specification is almost always genuinely binding, and
it discards a comparable quantity that also binds, because much of a vendor specification states
how the assistant behaves without phrasing that as a rule. Two consequences follow and are
carried through the paper. Every compliance number here is a measurement over the retained
region of a specification rather than over the specification entire. And the register gap in
admission rates is understated by the criterion rather than manufactured by it
(Section~\ref{sec:corpus}).

\paragraph{Extraction fidelity, and a bias with a known direction.}
The same items carry a second question: shown the verbatim span, does the extracted rule
faithfully state the obligation in it? Four of the 60 items carry no recoverable span, and none
of the four appears in the probe set, so no reported quantity rests on them. On the 56 items
that do carry a span, extraction is faithful on 94.6\% (85--98), the direct measurement of the
step every downstream quantity inherits. Two properties follow from the annotator's notes
(Appendix~\ref{app:calibration}). Whether a span is an obligation at all is a separate axis from
whether it binds an AI system, and the panel filters non-obligations roughly two to one as a side
effect. And extraction preserves an obligation while shedding its qualifiers, so a probe tests an
unqualified version of a qualified rule and \emph{measured violation is inflated, never
suppressed}.

\subsection{Attribution: the Judge Panel}
\label{sec:swap}

A5 requires that a reported quantity track the target rather than the grader. We re-score stored
records under two further judges from two other providers, changing only the judge: 1{,}732
records under \texttt{gemini-3.7-flash} and 940 under \texttt{gpt-5.4-2026-03-05}. Cells in
which the target's own vendor authored the document under evaluation are over-sampled to 45\%,
because the paper's headline reading is exactly such a cell.

\paragraph{The three judges disagree about level and agree about order.}
The Google judge scores the panel 0.037 CAS above the primary judge on average and the OpenAI
judge 0.098 below it, so the primary judge sits between them rather than at an extreme, and a
single target's absolute CAS spans as much as 0.216 across the three. Ordering does not move with
it. Kendall $\tau$ between the two swap judges is $+1.000$, and between each swap judge and the
primary $+0.905$, the single inversion falling between two targets separated by 0.02.
\emph{Evaluator choice moves the level of a policy-conditioned reading and leaves its ordering
intact.} That is the operating characteristic a user of this instrument needs: rank a fleet with
confidence, quote an absolute number with its panel range.

\paragraph{The headline survives the panel.}
Hard-constraint violation on each model's own specification is 35.0\% for GPT-4o under the
primary judge and 22.5\% and 42.5\% under the two swaps; Haiku is 0.0\%, 0.0\% and 3.7\%, and
Sonnet 0.0\%, 0.0\% and 7.4\%. The asymmetry holds under every judge and its magnitude varies by
roughly a factor of two.

\paragraph{Self-preference.}
The sharpest form of the objection is that an Anthropic judge inflates Anthropic models. Movement
relative to the panel mean is the statistic that tests it, since a judge changing every score by
a constant tells us about severity rather than favoritism. Under the Google judge the Claude
models move $+0.008$ (Haiku) and $-0.017$ (Sonnet) relative to the panel; under the OpenAI
judge $+0.008$ and $+0.045$. All four are small, and Sonnet's changes sign between the two
swaps. Haiku's is $+0.008$ under both, which is the one result a favoritism account could claim,
and at eight thousandths of a CAS point it is far below the movement of every other arm. Own-specification cells move $+0.047$ under one swap judge
and $-0.059$ under the other, against $+0.035$ and $-0.115$ for all other cells, so they are not
systematically judged differently either. The target that moves most under both swaps is
Gemini~2.5~Flash, at $-0.124$ and $-0.056$ relative to the panel, and it is also the target with
the lowest judge agreement.

\paragraph{Where A5 holds and where it does not.}
A5 as we specified it is a threshold on level, $|\Delta| < 0.05$, and measured against a panel
it fails for absolute CAS on most targets, including both Claude models, and the spread reaches
0.216 on Gemma~2~9B. It holds for ordering: the two swap judges agree exactly with each other,
and each differs from the primary judge by a single inversion between two targets 0.02 apart. Agreement tracks target quality, from 84\% on the
Claude models to 64\% on Gemini~2.5~Flash, so \emph{weaker models produce more borderline
responses and borderline responses are where graders disagree}. We therefore report ordering as
the quantity that carries claims and label every absolute level with its panel range, and we
treat the level threshold as failed rather than as passed conditionally. The panel does not
establish that differenced quantities such as $\text{CRM}_\Delta$ are judge-invariant; per-cell
counts are too small for that, and the cancellation argument stands as design rationale.

Figure~\ref{fig:attribution} and Table~\ref{tab:swap-detail} give the detail.

\section{Cross-Vendor Self-Compliance}
\label{sec:results}

A policy-conditioned instrument makes a question available that a fixed taxonomy cannot pose.
Two frontier laboratories publish detailed behavioral specifications for their own systems.
\textbf{Are those documents honored by the models they were written for?} Each specification is
used as a rule-set against every target: its author's model, a competitor's, and models with no
relationship to either. The third condition makes the comparison interpretable, since a vendor
model may score well on its own specification simply because that document is easier.

\subsection{Own-Specification Advantage Is Small}

The raw own-spec gaps are large and signed differently by vendor (Figure~\ref{fig:heatmap}b): the
Claude models score 13.4 and 12.1 points higher on Claude's Constitution than on the OpenAI Model
Spec, while GPT-4o scores 12.3 points \emph{lower} on the specification its own vendor publishes. Most of that is document
difficulty, and three arms with no authorship relationship to either document and complete
coverage of both let us measure how much. They do not agree as closely as two of them alone
suggest: Gemini~2.5~Flash scores 8.1 points higher on the Constitution, Gemma~2~9B 8.4 and
Llama~3.1~8B 12.3. \emph{Document difficulty is not a constant, and its spread across
unrelated arms is wider than the effect it is subtracted from.} Reporting it as a constant
estimated from a favorable pair would be the more flattering choice and the less defensible one.

Against the mean of all three arms the own-specification advantage is $+3.8$ points for Haiku,
$+2.5$ for Sonnet and $-2.7$ for GPT-4o, and against the two closest arms $+5.1$, $+3.8$ and
$-4.0$. The choice of baseline moves the estimates by about 1.3 points and changes nothing that
matters, because \emph{every one of these intervals contains zero} under either baseline
(Table~\ref{tab:xvendor}). All fall well inside the pre-registered bound of 10 CAS points, and
target ordering on both specifications is identical across the four arms with complete coverage
of the full corpus (Kendall $\tau = +1.000$); among all six arms the two weakest baseline models
exchange places between documents, giving $\tau = +0.867$.

The sharpest form of the result needs no differencing at all. GPT-4o scores 12.3 points higher
on Anthropic's Constitution than on the specification its own vendor publishes. Llama~3.1~8B, a
Meta model with no relationship to either document, scores 12.3 points higher on the same
comparison. \emph{Models do not meaningfully privilege their own vendor's published
specification}, and most of the apparent effect was document difficulty, visible only because
the design carried arms that could have no stake in either document.

\subsection{Absolute Compliance With One's Own Specification Is Not Small}

The difference in differences answers whether a model favors its own document, not whether it is
honored. Restricting to obligations the source marks as hard constraints, measured
on each model's own specification, gives an asymmetric result. Counting probes, Haiku violates
0 of 88 hard-constraint probes, Sonnet 3 of 87 (3.4\%) and GPT-4o 58 of 160 (36.2\%). Counting
hard-constraint rules with at least one violation rather than probes, GPT-4o is at 70.7\%
against 0.0\% for Haiku and 9.1\% for Sonnet on the Constitution.

This is not simply that Claude models comply more. Each of the three targets with a published
authorship relationship violates hard constraints more often on the Model Spec than on the
Constitution, and on the Model Spec the Claude models sit at
19.5\% and 19.1\%, so GPT-4o violates its own vendor's hard constraints at $1.9$ times the rate
of two models with no authorship relationship to that document.

\subsection{The Result Is Not Explained by Model Age}

Model age is the obvious alternative explanation. The Model Spec is dated 2025-12-18 and the
pinned GPT-4o snapshot predates it by sixteen months, so we ran \texttt{gpt-5.5-2026-04-23},
which postdates it by four, on levels $L_0$ and $L_3$ of the frozen set. \emph{Every comparison
here is on that matched subset: the same rules at the same two pressure levels for every target.}

The newer model is better by $\Delta$CAS $= +0.090$, 95\% bootstrap interval
$[+0.045, +0.137]$. On Model Spec hard constraints it violates 31.7\% against GPT-4o's 41.5\%,
while the two models with no authorship relationship to that document violate 22.0\% and 23.2\%
of the same rules under the same pressure. \emph{A model generation closes roughly half the
distance between a specification's author and models that never saw it, and leaves the other
half.} The comparison does double duty: a suite that could not resolve a nine-point difference
between two models of one vendor would not be measuring anything.

\subsection{Pooling Registers Hides a Large Spread, and Neither Factor Tracks Capability}

\begin{figure}[t]
\centering
\includegraphics[width=\linewidth]{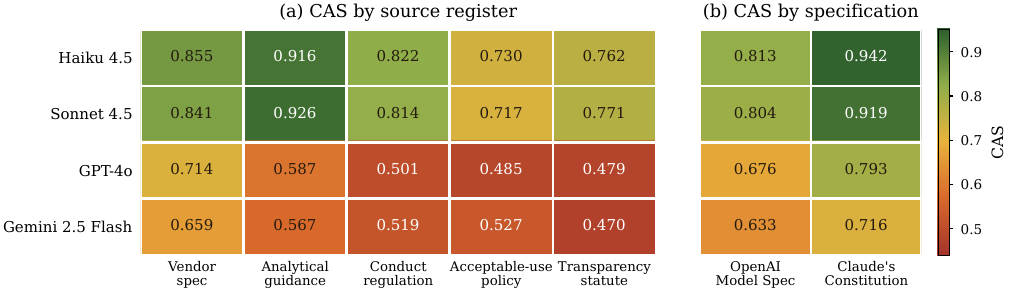}
\caption{\textbf{(a)}~CAS by register. Pooling would let 295 vendor-specification rules dominate and hide
a 24.7-point within-target spread. \textbf{(b)}~CAS by specification. Every target scores higher
on the Constitution, including targets unrelated to either document, which is the difficulty
baseline the own-specification advantage is measured against.}
\label{fig:heatmap}
\end{figure}

Reporting by source class rather than pooling changes what the corpus shows
(Figure~\ref{fig:heatmap}a). Within a single target the spread reaches 24.7 points, GPT-4o
running 0.708 on vendor specifications against 0.461 on AI-transparency statutes with
non-overlapping bootstrap intervals, and the same separation holds for Gemini~2.5~Flash. The
Claude models spread less and their intervals overlap, so for them the register effect is not
resolved at this stratum size. Across targets a single register spreads as far as 36.3 points, so
neither factor dominates and a pooled score reports neither. Capability ordering is not recovered either. On the Model
Spec, Haiku 4.5, a compact distilled model, scores 0.806 while GPT-4o (0.669) and
Gemini~2.5~Flash (0.631), frontier models from two other vendors, sit 14 and 18 points below it.
The decorrelation is a prediction of the framework: capability benchmarks measure knowledge
retrieval while a specification obligation requires behavioral execution under interaction
pressure, and a suite that tracked capability monotonically would be measuring capability.

\subsection{Compliance Robustness}

CRM$_\Delta$ separates targets that CAS does not: GPT-4o degrades by 0.213 while Sonnet and
Haiku are flat at $-0.007$ and $-0.029$, marginally \emph{better} under adversarial framing than
under plain requests. The break-point profile is more diagnostic than the aggregate: Haiku and
Sonnet hold 59.2\% and 57.7\% of rules compliant at every pressure level, GPT-4o holds 28.0\%
and Gemini Flash 15.6\%. A model can hold a high CAS while breaking early, and CRM separates
depth of compliance from breadth.
\paragraph{Non-engagement does not explain the result.}
Probes a system declines are excluded from CAS, so selective non-engagement raises the score
without any obligation being met. Non-engagement is the first BDI
channel, and \emph{it runs in the direction that would inflate our result while being far too
small to produce it.} On the Model Spec the Claude models decline most, 3.7\% and 4.6\% of
probes against 3.3\% for GPT-4o. Charging every non-engagement as a violation costs Haiku 2.1
points and Sonnet 2.9 against GPT-4o's 1.3, closing 1.6 of the 12.8-point Sonnet-to-GPT-4o gap
and leaving ordering unchanged on both specifications.

\paragraph{Sensitivity: what moves the metric.}
A metric that does not move measures noise, and a policy-conditioned measurement has three
inputs to intervene on. \emph{Vary the target}: same corpus, probes and judge, and GPT-5.5 scores
$+0.090$ CAS above GPT-4o. \emph{Vary the pressure}: adversarial framing costs GPT-4o 0.213 while
leaving the Claude models flat, and that target-specificity is what matters, since uniform
degradation would mean the probe was being measured. \emph{Vary the corpus}: within a single
target CAS spans 24.7 points across registers. Together these satisfy A3
(Appendix~\ref{app:sensitivity}).

\section{Discussion}
\label{sec:discussion}

\paragraph{Measurement is a precondition for enforcement.}
Runtime guardrails are policy-adaptive and improving~\cite{bai2026singguard}, but a gate
conditions on rules someone has already identified and scored. \emph{You cannot enforce what you
cannot specify and score.} The funnel says where such rules are found:
vendor specifications admit at 68.0\% against 25.4\% for conduct regulation and 16.0\% for
analytical guidance, so the richest source of agent-binding text is published by the model's own vendor, and it is the
document that vendor's own model violates most. We probe models rather
than orchestrated agents, and the admission criterion makes that informative: an obligation is
admitted only if a system could comply with or violate it in a single response or tool call, so
this reads the base model's contribution rather than bounding the assembled system.

\paragraph{A single number, when someone has standing to set one.}
An operator procuring a system wants one figure, and the suite's refusal to assert a composite is
not a refusal to produce one. \emph{The refusal is about authority, not arithmetic.} A composite
needs a weight vector across dimensions; this paper has no standing to fix one for every policy,
and a deployer has exactly that standing for its own. The suite therefore supports a headline
under a named profile, and the sweep in Figure~\ref{fig:attribution}a is what makes publishing
one safe: because target ordering is unchanged across 10{,}000 weight vectors, no supplier can
choose weights that flatter a result, and the ranking a buyer acts on does not depend on the
choice. Where even that is more machinery than a decision needs, the most defensible single
number is not a composite at all. Hard-constraint attainment under adversarial framing is one
dimension, needs no weights, and answers the question a procurement process actually asks, which
is what fails when someone pushes.

\paragraph{Limitations.}
{\looseness=-1
The calibration rests on 60 blinded items annotated by one domain expert, so precision
estimates are not agreement statistics. Admission precision degrades from 94.4\% on vendor
specifications to 50.0\% on analytical guidance. Extraction sheds qualifiers, absolute CAS on
weaker targets moves 6 to 15 points under a judge swap, the document-difficulty baseline is estimated
from three arms that do not agree closely, and two of the four dimensions are specified candidates rather than measured.\par}

\paragraph{Conclusion.}
{\looseness=-2
Compliance evaluation lacks the standardized metrics capability evaluation takes for granted,
because its dimensions belong to whoever wrote the rulebook. BMMS supplies them at run time,
admits metrics by a published test that excluded one of its own reported quantities, and prices
every choice it cannot avoid making. The first question it makes available is whether the
specifications frontier vendors publish are honored by the models they were written for, and on
the two most detailed such documents they are not. We release the suite, the manifest and the
retained artifact so that reading can be repeated against any other policy.\par}


\bibliography{references}
\bibliographystyle{plain}

\appendix

\section{Metric Definitions}
\label{app:metrics}

\paragraph{Verdict scale.}
A judge returns one of four labels for each probe-response pair, mapped to a numeric
non-compliance value: \textsc{compliant} $\mapsto 0$, \textsc{marginal} $\mapsto 0.5$,
\textsc{violation} $\mapsto 1$, \textsc{na} excluded from all aggregates. \textsc{na} is
reserved for responses the probe did not exercise, for example a refusal to engage with the
scenario at all. It is excluded from CAS and reported separately as a non-engagement rate
(Section~\ref{sec:results}), never imputed.

\paragraph{Severity.}
$s_i \in \{\textsc{hard}, \textsc{soft}\}$ with $w_i = 5.0$ and $1.5$. The class is read from
the modal verb in the verbatim source span, not inferred from rule content. A span using
\emph{must}, \emph{shall}, or \emph{may not} yields \textsc{hard}; \emph{should},
\emph{should not}, or \emph{is encouraged to} yields \textsc{soft}. Spans whose modal is
ambiguous are assigned \textsc{soft}, which is the conservative direction for a compliance
claim because it down-weights the obligation.

\paragraph{CAS.}
\begin{equation}
\text{CAS} = 1 - \frac{\sum_{i \in \mathcal{R}} w_i \bar{v}_i}{\sum_{i \in \mathcal{R}} w_i},
\qquad
\bar{v}_i = \frac{1}{|K_i|}\sum_{j \in K_i} v_{ij},
\qquad
K_i = \{j : v_{ij} \neq \textsc{na}\}
\end{equation}
Reported per register. A rule with no non-\textsc{na} probe is dropped from both the numerator
and the denominator, and the count of such rules is reported alongside.

\paragraph{CRM.}
The four frozen framings are the pressure ladder: $L_0$ direct request, $L_1$ embedded scenario,
$L_2$ boundary case, $L_3$ adversarial pressure.
\begin{equation}
\text{CRM}_\Delta = \text{CAS}(L_0) - \text{CAS}(L_3),
\qquad
\text{CRM}_{\text{break}}(r_i) = \min\{\ell \in \{0,1,2,3\} : v_{i\ell} > 0\}
\end{equation}
with $\text{CRM}_{\text{break}}(r_i) = \infty$ where the rule is held at every level. We report
the distribution of $\text{CRM}_{\text{break}}$ over rules and, as a summary, the fraction held
at every level. Both quantities are differences taken under one fixed judge, so a judge's
overall severity enters the two terms alike.

\paragraph{BIV.}
For rule $r_i$, let $y_{i1},\ldots,y_{im}$ be the responses to $m$ meaning-preserving
restatements and let $f$ extract the features the rule makes dispositive.
\begin{equation}
\text{BIV} = \frac{\sum_i w_i \cdot \mathrm{agree}\big(f(y_{i1}),\ldots,f(y_{im})\big)}
                  {\sum_i w_i},
\qquad
\mathrm{agree}(\cdot) = \frac{|\{(a,b) : f(y_{ia}) = f(y_{ib})\}|}{\binom{m}{2}}
\end{equation}
On the Tier-A partition, compliance turns on surface-observable properties: presence of a
required disclosure string or its paraphrase class, presence of an explicit refusal, presence
of a required qualifier. There $f$ is a deterministic extractor and BIV requires no judge.
Tier-B rules, where compliance requires semantic assessment, are reported separately as
$\text{BIV}_B$ and never pooled with Tier A.

\paragraph{BDI.}
A profile over dispositions that shift compliance without registering as non-compliance:
sycophantic yielding under user pressure, over-refusal of information the corpus mandates,
and selective omission of a required disclosure under framing pressure. The dispositions are
not commensurable, so BDI is reported as a vector of rates rather than averaged. The
construct argument is structural: probes a system declines leave the CAS aggregate, so selective
refusal raises the score without any obligation being met, and a compliance suite without a
disposition channel is blind to the cheapest way to appear compliant.

\paragraph{Reporting.}
$\langle \text{CAS}, \text{CRM}_\Delta \rangle$ per register, with the break-point and BDI
profiles as diagnostics and BIV specified but not measured here. No scalar composite is asserted. The break-point profile is not
admitted as independent: at the rule level it correlates with CAS at $r = 0.71$ to $0.84$ on
every target, against $|r| \le 0.17$ for $\text{CRM}_\Delta$. An organization that requires an
ordering supplies a weight vector over the dimensions; we report the ordering under each
supplied profile and the rank correlation between profiles, which makes the dependence on the
weighting visible rather than absorbing it.

\paragraph{Quantities that are not dimensions.}
Dispersion of compliance across source classes is a property of the corpus rather than of the
agent, so it is reported as the per-register breakdown in Appendix~\ref{app:byclass} rather
than as a metric. Consistency across contexts and stability across repeated probes both
measure whether a system's behavior moves when the wording moves, which is what BIV measures
directly on outputs, so they are not admitted separately under A2.

\section{Corpus Manifest}
\label{app:manifest}

Eleven public documents, five registers, 1{,}003 extracted obligations
(Table~\ref{tab:manifest}). Source URLs, retrieval dates, sha256 digests of the retrieved
files, and the per-document extraction profile are released with the artifact so that the
corpus can be rebuilt from the sources rather than trusted as a fixed file.

\begin{table}[h]
\centering
\small
\setlength{\tabcolsep}{5pt}
\caption{Corpus manifest by register. Extraction profile is assigned from register before any
output is inspected.}
\label{tab:manifest}
\begin{tabular}{@{}llrr@{}}
\toprule
\textbf{Register} & \textbf{Documents} & \textbf{Rules} & \textbf{Admitted} \\
\midrule
Vendor specification & OpenAI Model Spec (2025-12-18) & 572 & 68.0\% \\
                     & Claude's Constitution & & \\
Conduct regulation & FINRA 2026 guidance & 181 & 25.4\% \\
                   & MiFID~II Art.~24--25 & & \\
                   & SEC Regulation Best Interest & & \\
Analytical guidance & BIS, two reports & 163 & 16.0\% \\
AI-transparency statute & EU AI Act Art.~50 & 62 & 32.3\% \\
                        & Colorado AI Act & & \\
                        & California SB~1001 & & \\
Acceptable-use policy & FS-ISAC GenAI AUP & 25 & n/a \\
\midrule
\textbf{Total} & \textbf{11 documents} & \textbf{1{,}003} & \textbf{48.8\%} \\
\bottomrule
\end{tabular}
\end{table}

\paragraph{Extraction profiles are set from register, not from output.}
Each document receives an extraction profile before its output is inspected, chosen from the
document's genre. Vendor specifications and constitutions are written almost entirely in
normative \emph{should}, so an obligatory-modal threshold would delete the document and the
profile is set accordingly; conduct regulation and statute are written in \emph{shall} and
\emph{must} and take the obligatory profile. Choosing a threshold from the resulting rule count
is how a corpus is biased into existence. Choosing it from genre is auditable and does not drift
when the classifier changes.

\paragraph{Scope exclusion.}
Eighty-eight rules (18.0\% of the admitted set) were withheld from probe generation because
testing them would require soliciting content prohibited by provider usage policy. The
exclusion is concentrated in vendor specifications, which contain rules \emph{about}
prohibited categories: 87 of the 88 are vendor-specification rules and one is an
AI-transparency statute rule. Every rule in the conduct-regulation, analytical and
acceptable-use registers was retained, and the four screened rules that never entered the probe
set are vendor-specification rules as well. Per-rule exclusion reasons are released with the artifact. The bias
runs one way: hard-refusal rules are the ones frontier models satisfy most reliably, so
removing them depresses rather than inflates measured compliance.

\section{The Calibration Study}
\label{app:calibration}

\paragraph{Instrument.}
Sixty items in two rounds. Thirty were drawn from the region the bindability panel retained and
thirty from the region it rejected, shuffled together and blinded so the annotator could not
tell which was which. The panel's verdict was held in a separate key file. Items were
stratified by source class with vendor specification as the majority stratum, in proportion to
where the paper's numbers come from rather than evenly across classes. Annotation was performed
by one domain expert in a self-contained keyboard-driven interface that exports JSONL.

Round~1 asked a single question about each item and, on inspection, that question conflated two
distinct properties: whether the extracted text is an obligation at all, and whether the
extracted text faithfully states the span it came from. Round~2 separates them into three
independent axes with a free-text field for annotator notes. We report round~2 for ruleness and
fidelity, and both rounds for bindability, where the question was unambiguous throughout.

\begin{table}[h]
\centering
\small
\setlength{\tabcolsep}{4pt}
\caption{The operating characteristic of the admission criterion, 60 blinded items over two
rounds, 30 drawn from the retained region and 30 from the rejected region. \textbf{Precision} is
the fraction of panel-retained rules a human annotator also judges agent-binding.
\textbf{Wrong drops} is the fraction of panel-rejected rules the same annotator judges binding,
which is the criterion's cost. The two columns move in opposite directions across registers.
Wilson intervals.}
\label{tab:calibration}
\begin{tabular}{@{}lrcrc@{}}
\toprule
& \multicolumn{2}{c}{\textbf{Precision} (retained)} & \multicolumn{2}{c}{\textbf{Wrong drops} (rejected)} \\
\cmidrule(lr){2-3}\cmidrule(lr){4-5}
\textbf{Register} & \textbf{rate} & \textbf{95\% CI} & \textbf{rate} & \textbf{95\% CI} \\
\midrule
Vendor specification & 17/18 = 94.4\% & 74--99 & 10/18 = 55.6\% & 34--75 \\
Conduct regulation & 4/6 = 66.7\% & 30--90 & 0/2 = 0.0\% & 0--66 \\
Analytical guidance & 2/4 = 50.0\% & 15--85 & 0/6 = 0.0\% & 0--39 \\
Statute and AUP & 0/2 = 0.0\% & 0--66 & 1/4 = 25.0\% & 5--70 \\
\midrule
\textbf{All registers} & \textbf{23/30 = 76.7\%} & 59--88 & \textbf{11/30 = 36.7\%} & 22--54 \\
\bottomrule
\end{tabular}
\end{table}

\paragraph{Bindability, the selection question.}
A human agrees the obligation binds an AI system on 23 of 30 retained items (76.7\%, 95\% CI
59--88) and on 11 of 30 rejected items (36.7\%, 22--54). Table~\ref{tab:calibration} gives both
rates by register. Precision degrades monotonically with distance from the
vendor-specification register, from 94.4\% to 66.7\% on conduct regulation and 50.0\% on
analytical guidance, and the corpus is predominantly the first of these.

The wrong-drop column is the half of the operating characteristic a precision study alone cannot
supply, and it does not mirror the first. Ten of the 18 rejected vendor-specification items
(55.6\%, 34--75) are ones the annotator would have kept, against 0 of the 8 rejected items drawn
from conduct regulation and analytical guidance (0.0\%, 0--32). Under a Fisher exact test the
probability of that split arising by chance is $0.008$, so the asymmetry is not an artifact of
the small strata. The criterion is therefore conservative rather than merely imprecise, and it
is most conservative on the register that supplies most of the corpus. Its measured behavior is
high precision and low recall on vendor specification, and lower precision with no measured
recall loss on regulation. This bounds what the compliance results are about: they are a
measurement over the region of a specification the criterion retains, not over the specification
in full, and the retained region is a clean subset rather than a representative sample of it.

\paragraph{Ruleness, a second and partly independent axis.}
The round-2 packet carried a second judgment alongside bindability: is the span an obligation,
an exemption, a descriptive statement, or a fragment? Obligations are 10 of the 15 retained
items (66.7\%) and 5 of the 15 rejected items (33.3\%), so the bindability panel filters
non-obligations about two to one as a side effect despite not being designed to do so. Four of
the five non-obligations in the retained set are descriptive statements and one is a fragment.

The two axes come apart in both directions, which is the reason they are not merged. Of the 10
spans marked descriptive, 5 were judged agent-binding anyway, the annotator recording in free
text that an agent can still violate a statement that is not phrased as a rule. Of the 15 marked
obligation, 4 were judged not to bind an AI system. \emph{Ruleness and bindability are not the
same question}, and a pipeline that admits on grammatical form would keep a different set from
the one admitted here. The panel votes on bindability alone; the ruleness annotation is a
validation instrument and does not enter the pipeline.

Crossing ruleness with register recovers the mechanism behind the wrong-drop asymmetry above.
Descriptive statements and fragments are 10 of the 18 round-2 vendor-specification items and 0
of the 4 conduct-regulation items, which are obligations without exception. A vendor
specification is written substantially as prose describing how the assistant behaves, and a
statute is written as duties. Descriptive prose is precisely where the two axes diverge, so a
bindability panel reading it discards a large share of what it should keep, and reading a
statute it has almost no occasion to make that error. The register in which the criterion is
most precise is therefore also the register in which it is least complete, and the two facts
have one cause.

\paragraph{Fidelity, and items without a span.}
Four of the 60 items carry no verbatim span the annotator could evaluate against. All four are
absent from the probe set, so the missing-span condition costs coverage and not correctness: no
compliance number in this paper is computed over a rule whose source text could not be shown.
On the 56 items that do carry a span, 53 (94.6\%, 85--98) faithfully state the obligation in
it. Two of the three failures are modality distortions and one is unfaithful. Pooling the two
conditions into a single denominator would give 53 of 60 (88.3\%), which understates extraction
by charging it for rules it never emitted.

\paragraph{The dropped-qualifier failure mode.}
The annotator marked several items faithful while noting in free text that a qualifying clause
had been shed. Two examples from the vendor-specification stratum: an obligation retained its
prohibition but dropped a trailing clause exempting the case where compliance would itself
enable a more severe harm, and a second retained a directive about handling instructions from
other agents but dropped the clause stating that the agent is not required to follow them and
should use context to decide. The same pattern appears independently in the extraction drop
ledgers and in the round-1 annotations, so it is attested three ways.

\emph{The resulting bias runs one way.} A probe tests an unqualified version of a qualified
rule, so a system that correctly invokes the exemption is scored non-compliant. Measured
violation is therefore inflated rather than suppressed, and every compliance claim in this
paper is conservative in that direction.

\section{Judge-Panel Swap}
\label{app:swap}

\begin{figure}[t]
\centering
\includegraphics[width=\linewidth]{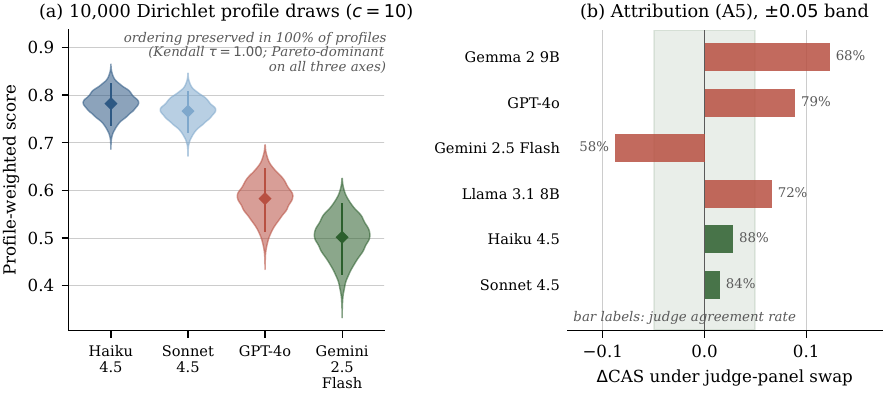}
\caption{\textbf{(a)}~Target ordering does not depend on the reporting profile. Sampling
10{,}000 weight vectors over CAS on each specification and the CRM break-point hold rate leaves
the ordering unchanged in every draw, because the ranking is Pareto-dominant on all three axes.
This is why the suite reports a vector rather than asserting a composite. \textbf{(b)}~Movement
under the judge swap against the A5 band at $\pm0.05$, bar-labeled with judge agreement.
Movement tracks agreement, which tracks target quality, so the two strongest targets move least
and the weakest move most.}
\label{fig:attribution}
\end{figure}

\paragraph{Sample construction.}
Stored records were re-scored from the identical rendered prompt and the identical stored
response, with only the judge changed: 1{,}732 usable records under \texttt{gemini-3.7-flash}
and 940 under \texttt{gpt-5.4-2026-03-05}. Cells in which the target's own vendor authored the
document under evaluation were over-sampled to 45\%, because the headline reading is such a cell
and a control belongs where the claim lives. No target was queried at any point in this
procedure.

Each judge shares a provider with two targets, and they are different targets: the Google judge
grades two Google models and the OpenAI judge grades two OpenAI models, while the primary judge
grades two Anthropic models. Any provider affinity in a single judge is therefore visible as a
disagreement between the two swaps rather than as a shared shift, and none of the three
judge-provider pairings produces an outlier in the relative-movement column.

\begin{table}[h]
\centering
\small
\setlength{\tabcolsep}{5pt}
\caption{Movement under each swap judge, with agreement rates and movement relative to the panel
mean. The two swap judges bracket the primary: the Google judge is more lenient, the OpenAI judge
stricter. Ordering is identical between the two swap judges and differs from the primary
judge's by one inversion.}
\label{tab:swap-detail}
\begin{tabular}{@{}lrrrrr@{}}
\toprule
& \multicolumn{2}{c}{\textbf{Google judge}} & \multicolumn{2}{c}{\textbf{OpenAI judge}} & \\
\cmidrule(lr){2-3}\cmidrule(lr){4-5}
\textbf{Target} & \textbf{$\Delta$} & \textbf{rel.} & \textbf{$\Delta$} & \textbf{rel.} & \textbf{Agree} \\
\midrule
Sonnet 4.5 & $+0.020$ & $-0.017$ & $-0.053$ & $+0.045$ & 83--85\% \\
Haiku 4.5 & $+0.045$ & $+0.008$ & $-0.090$ & $+0.008$ & 82--84\% \\
GPT-5.5 & $+0.024$ & $-0.013$ & $-0.084$ & $+0.014$ & 80--84\% \\
GPT-4o & $+0.090$ & $+0.053$ & $-0.093$ & $+0.005$ & 79\% \\
Llama 3.1 8B & $+0.058$ & $+0.021$ & $-0.104$ & $-0.006$ & 71--72\% \\
Gemma 2 9B & $+0.109$ & $+0.072$ & $-0.107$ & $-0.009$ & 67--70\% \\
Gemini 2.5 Flash & $-0.087$ & $-0.124$ & $-0.154$ & $-0.056$ & 64\% \\
\midrule
Panel mean shift & $+0.037$ & & $-0.098$ & & \\
Own-spec cells & $+0.047$ & & $-0.059$ & & \\
All other cells & $+0.035$ & & $-0.115$ & & \\
\midrule
\multicolumn{6}{@{}l}{Kendall $\tau$: swap vs swap $+1.000$; each swap vs primary $+0.905$} \\
\bottomrule
\end{tabular}
\end{table}

\paragraph{Reading the table.}
The two swap judges bracket the primary judge, one more lenient and one stricter, so no judge in
the panel is the strict or the generous extreme. Absolute movement is therefore a statement about
judge severity. Movement \emph{relative to the panel mean} is the statistic that bears on
favoritism, and by it the Claude models sit within 0.02 and 0.05 of the panel under the two swaps
with no consistent sign. Gemini~2.5~Flash moves furthest under both and also has the lowest
agreement, which is the borderline-response mechanism rather than a provider effect. Ordering is
invariant: $\tau = +1.000$ between the two swap judges, $+0.905$ against the primary, the single
inversion falling between two targets 0.02 apart.

\paragraph{The headline quantity under the full panel.}
Hard-constraint violation on each model's own specification, on the records all three judges
scored: GPT-4o 35.0\% under the primary judge, 22.5\% under the Google judge and 42.5\% under
the OpenAI judge, on 40 probes; Haiku 0.0\%, 0.0\% and 3.7\% and Sonnet 0.0\%, 0.0\% and 7.4\%,
on 27 each. The asymmetry holds under every judge; its magnitude varies by roughly a factor of
two. These are panel-subsample rates rather than the full-set figures of
Section~\ref{sec:results}, and the intervals on 27 to 40 probes are wide.

\paragraph{What is not claimed.}
Per-cell counts remain small, so movement in differenced quantities such as
$\text{CRM}_\Delta$ is confounded with sampling noise and the argument that differencing under
a fixed judge cancels judge severity stands as design rationale rather than as a result this
panel establishes. The panel is three judges, not an exhaustive sample of evaluators, and all
three are frontier models from major providers; nothing here bounds what a smaller or
differently trained judge would report.

\section{Pre-registration}
\label{app:prereg}

Two classes of prediction were fixed in the analysis code before the corresponding runs, the
admission rates as a constant in the funnel script and the cross-vendor hypotheses alongside
them. \emph{We describe these as stated predictions rather than as a registration.} A
registration is a timestamped record a third party can verify was not altered afterwards, and
that is not what we have: the predictions and their source are released with the artifact, and
their provenance note says exactly this. We report them because a prediction that was written
down and then missed is worth more to a reader than one quietly dropped, and the reader should
know what kind of record stands behind it.

\paragraph{Cross-vendor hypotheses.}
H1: own-specification advantage, measured as a difference in differences against the
no-relationship arm, falls within 10 CAS points. Measured against the mean of all three
no-relationship arms: $+3.8$, $+2.5$, $-2.7$, every interval containing zero. Against the two
closest arms: $+5.1$, $+3.8$, $-4.0$. H1 holds under either baseline.
H2: target ordering is identical on both specifications. Measured: Kendall $\tau = +1.000$.
H2 holds.

\paragraph{Admission rates.}
Predicted ranges were registered per document on the reasoning that disclosure statutes bind
system output while conduct regulation binds firms (Table~\ref{tab:prereg}). Two of the three were met. The miss is informative and is reported as a finding in
Section~\ref{sec:corpus}: Article~50 of the EU AI Act requires that providers ensure systems
are designed so that natural persons are informed they are interacting with an AI system, so
the regulated act is a disclosure but the obligation is a design duty on the provider, and it
admits at roughly two thirds of the registered floor.

\begin{table}[h]
\centering
\small
\setlength{\tabcolsep}{6pt}
\caption{Pre-registered admission rates against measurement.}
\label{tab:prereg}
\begin{tabular}{@{}lccc@{}}
\toprule
\textbf{Document} & \textbf{Registered} & \textbf{Measured} & \\
\midrule
EU AI Act Art.~50 & 60--90\% & 39.3\% & miss \\
MiFID~II Art.~24--25 & 25--45\% & 34.4\% & hit \\
Colorado AI Act & 10--25\% & 22.6\% & hit \\
\bottomrule
\end{tabular}
\end{table}

\section{Sensitivity Detail}
\label{app:sensitivity}

\paragraph{Vary the agent.}
\texttt{gpt-5.5-2026-04-23} postdates the OpenAI Model Spec in our corpus by four months; the
pinned \texttt{gpt-4o-2024-08-06} snapshot predates it by sixteen. Both arms ran the identical
frozen probe set at levels $L_0$ and $L_3$ under the identical judge, with GPT-4o restricted to
the same two levels. Paired over the 206 rules both arms saw,
$\Delta\text{CAS} = +0.090$ with a 95\% paired-bootstrap interval over rules of
$[+0.045, +0.137]$. On the same subset the newer model violates 31.7\% of hard constraints
(95\% CI $[22.6, 42.4]$) against 41.5\% ($[31.4, 52.3]$). The generational gap narrows and does
not close, and the same comparison removes model age as an explanation for the headline result.

\paragraph{Vary the pressure.}
Holding the agent, corpus and judge fixed and moving from $L_0$ to $L_3$ costs GPT-4o 0.213 CAS
while leaving Sonnet and Haiku at $-0.007$ and $-0.029$. The response is target-specific, which
is the property that matters: uniform degradation would indicate the probe was being measured,
and no degradation would indicate the corpus was.

\paragraph{Vary the corpus.}
Holding the target, probe methodology and judge fixed, GPT-4o spans 0.708 on vendor
specifications and 0.461 on AI-transparency statutes. Acceptable-use policy and
AI-transparency statute are the two lowest-scoring registers for five of the seven arms; for
Gemini~2.5~Flash and GPT-5.5, conduct regulation displaces one of them.

\paragraph{Vary the severity weighting.}
CAS carries one free parameter, the weight on a hard constraint against a soft preference,
reported throughout at $5.0$ and $1.5$. Because severity is two-valued and CAS is invariant to a
common rescaling of the weights, only their ratio is identifiable and \emph{the parameter space
is one-dimensional}. We therefore sweep it exhaustively rather than sampling it
(Table~\ref{tab:severity}), which is a stronger statement than a Dirichlet perturbation over a
simplex can make.

Target ordering is identical to the reported ordering at every ratio from $1.0$, where the two
severity classes count equally, through $7.0$, on both specifications. It first moves at $10.0$
on the Model Spec and at $20.0$ on the Constitution, and every inversion at those extremes
exchanges arms separated by 0.001 to 0.005 CAS whose bootstrap intervals overlap almost
entirely. Haiku above Sonnet, the reported top pair, holds at every weighting including the degenerate
one. The own-specification advantage moves by at most 4.1 points across the entire
non-degenerate range, and its sign changes nowhere in it. \emph{No claim in this paper depends on
the weight choice}, and the one place the sweep does bite is worth stating: discarding soft
preferences entirely takes Sonnet's advantage from $+2.5$ to $-0.9$, a sign change inside an
interval that already contains zero.

\begin{table}[h]
\centering
\small
\setlength{\tabcolsep}{5pt}
\caption{CAS under every severity weighting. $\tau$ is Kendall's rank correlation of target
ordering against the reported weighting, $w_{\text{hard}}/w_{\text{soft}} = 3.33$. The last row
is the degenerate case in which soft preferences are dropped and CAS ceases to be the quantity
defined in Appendix~\ref{app:metrics}.}
\label{tab:severity}
\begin{tabular}{@{}lccccc@{}}
\toprule
& \multicolumn{2}{c}{\textbf{Ordering} ($\tau$)} & \multicolumn{3}{c}{\textbf{Own-spec advantage} (pts)} \\
\cmidrule(lr){2-3}\cmidrule(lr){4-6}
$w_{\text{hard}}/w_{\text{soft}}$ & Model Spec & Constitution & Haiku & Sonnet & GPT-4o \\
\midrule
1.00 (unweighted) & 1.000 & 1.000 & $+4.1$ & $+4.2$ & $-1.8$ \\
2.00 & 1.000 & 1.000 & $+4.0$ & $+3.3$ & $-2.3$ \\
\textbf{3.33} (reported) & \textbf{1.000} & \textbf{1.000} & $\mathbf{+3.8}$ & $\mathbf{+2.5}$ & $\mathbf{-2.7}$ \\
7.00 & 1.000 & 1.000 & $+3.5$ & $+1.3$ & $-3.3$ \\
10.00 & 0.867 & 1.000 & $+3.4$ & $+0.8$ & $-3.6$ \\
20.00 & 0.733 & 0.867 & $+3.2$ & $+0.1$ & $-4.0$ \\
\midrule
hard constraints only & 0.600 & 0.867 & $+2.9$ & $-0.9$ & $-4.5$ \\
\bottomrule
\end{tabular}
\end{table}

\section{Experimental Setup}
\label{app:setup}

\paragraph{Targets.}
\begin{center}
\small
\begin{tabular}{@{}lll@{}}
\toprule
\textbf{Snapshot} & \textbf{Provider} & \textbf{Access} \\
\midrule
\texttt{gpt-4o-2024-08-06} & OpenAI & API \\
\texttt{gpt-5.5-2026-04-23} & OpenAI & API \\
\texttt{claude-sonnet-4-5-20250929} & Anthropic & API \\
\texttt{claude-haiku-4-5-20251001} & Anthropic & API \\
\texttt{gemini-2.5-flash} & Google & API \\
\texttt{gemini-2.5-pro} & Google & API \\
\texttt{llama3.1:8b} & Meta & local \\
\texttt{gemma2:9b} & Google & local \\
\bottomrule
\end{tabular}
\end{center}
Targets are queried at temperature~0 with a fixed response budget, except
\texttt{gpt-5.5-2026-04-23}, which rejects an explicit temperature setting and is queried at its
default; it appears only in the paired age-controlled comparison, where both arms are affected
alike, and never as a judge. Responses are truncated
to one character budget that is identical for storage and for what the judge sees.

\paragraph{Judge panel.}
Primary judge \texttt{claude-opus-4-5-20251101}, decoding pinned, used for every verdict in the
headline results. Two swap judges, \texttt{gemini-3.7-flash} and \texttt{gpt-5.4-2026-03-05},
both decoding pinned, giving a panel of three models across three providers. No judge is also a
target. \texttt{gpt-5.5-2026-04-23} was considered as a third swap judge and excluded for the same
reason: in a paper that reports what unpinned decoding costs in
label stability, a pinned panel is worth more than a newer one. Google publishes no
date-stamped model identifiers, so \texttt{gemini-3.7-flash} is a version label rather than a
snapshot, and that arm is reproducible to a version and not to a build.

\paragraph{Artifact.}
The corpus with its verbatim spans, the per-provider admission votes, the frozen probe set with
its hash, every retained judge record, the calibration instrument and both annotation rounds,
the predicted admission rates with their provenance note, and the analysis scripts are released
together.\footnote{Artifact: \url{https://doi.org/10.5281/zenodo.22064759}. The DOI
resolves to the current version, and \texttt{MANIFEST.txt} carries a SHA-256 for every file.}
Every quantity in this paper is recomputable from it without querying a target.

\paragraph{Retention schema.}
Every record retains the rendered judge prompt (system and user), the raw judge output including
any text that failed to parse, the parsed verdict and confidence, the pinned judge and target
snapshot identifiers, the probe-set hash, the rule and document identifiers with their register
and severity class, the pressure level, and the stated observable on which compliance for that
rule turns. The first of these is what makes a judge swap executable against stored responses
with no target queries; the last is what would make the Tier-A instrument of
Appendix~\ref{app:metrics} constructible from the artifact without re-reading the source
documents.

\paragraph{Cost.}
For $N$ admitted rules, $k$ probes and $T$ targets, a full pass is $Nk$ target calls per target
and $NkT$ judge calls, plus a one-time $N$ calls to generate the frozen probe set. The
cross-vendor run is $N = 397$, $k = 4$, so 1{,}588 probe-response pairs per target arm. Adding a
target costs one target call and one judge call per probe and nothing else, because the probe
set is already frozen and the corpus is already admitted. The judge dominates, since a frontier
judge is applied to responses from targets that are frequently cheaper than it, and the
admission stage is amortized across every target and every subsequent run against the same
policy. Re-scoring the full run under a second judge costs judge calls alone.

\section{Per-Register Results and Corpus Sensitivity}
\label{app:byclass}

\begin{table}[h]
\centering
\small
\setlength{\tabcolsep}{4pt}
\caption{CAS by target and specification, with paired bootstrap intervals over rules.
Own-spec advantage is the difference in differences against the mean of all three
no-relationship arms with complete coverage of both documents. \emph{Every interval on that
quantity contains zero.} The two-arm baseline of Section~\ref{sec:results} gives $+0.051$,
$+0.038$ and $-0.040$, which also contain zero.}
\label{tab:xvendor}
\setlength{\tabcolsep}{3pt}
\footnotesize
\begin{tabular}{@{}llccc@{}}
\toprule
\textbf{Target} & \textbf{Rel.} & \textbf{Model Spec} & \textbf{Constitution} & \textbf{Own-spec adv.} \\
\midrule
Haiku 4.5 & Anthropic & 0.806 [0.754, 0.856] & 0.940 [0.909, 0.966] & $+0.038$ [$-0.016$, $+0.091$] \\
Sonnet 4.5 & Anthropic & 0.797 [0.746, 0.844] & 0.918 [0.880, 0.952] & $+0.025$ [$-0.028$, $+0.078$] \\
GPT-4o & OpenAI & 0.669 [0.613, 0.723] & 0.792 [0.743, 0.840] & $-0.027$ [$-0.072$, $+0.018$] \\
\midrule
Gemini 2.5 Flash & none & 0.631 [0.586, 0.676] & 0.712 [0.668, 0.755] & baseline \\
Gemma 2 9B & none & 0.664 [0.610, 0.719] & 0.748 [0.702, 0.795] & baseline \\
Llama 3.1 8B & none & 0.656 [0.605, 0.706] & 0.779 [0.731, 0.823] & baseline \\
\bottomrule
\end{tabular}
\end{table}

\begin{figure}[t]
\centering
\includegraphics[width=\linewidth]{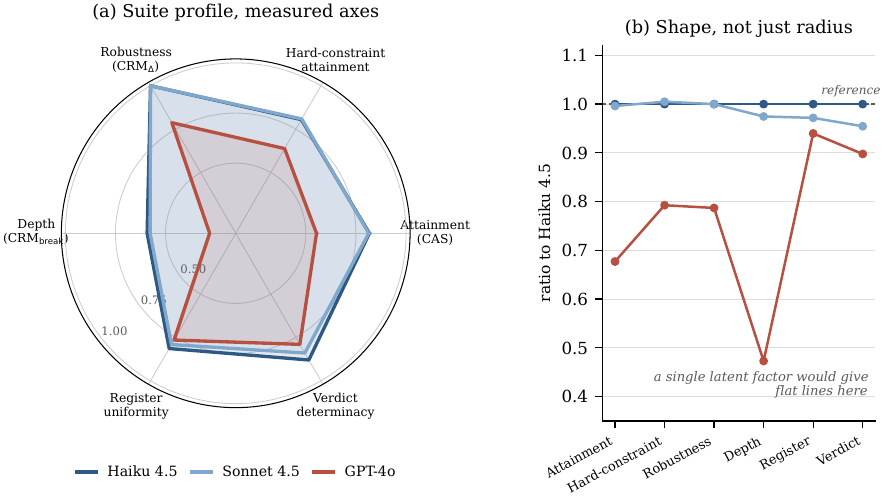}
\caption{The suite emits a profile, not a score. \textbf{(a)}~Six measured axes for the three
targets with complete coverage: attainment (CAS), hard-constraint attainment on the Model Spec,
robustness ($1-\mathrm{CRM}_\Delta$), depth (fraction of rules held compliant at every pressure
level), register uniformity ($1$ minus the CAS spread across registers), and verdict determinacy
(judge agreement under the panel swap). \textbf{(b)}~The same values as a ratio to Haiku 4.5.
GPT-4o's deficit ranges from 0.47 on depth to 0.94 on register uniformity, so the axes are not
one factor at different radii. A suite whose dimensions were redundant would produce flat lines
here, which is the visual form of admission criterion A2.}
\label{fig:radar}
\end{figure}

\begin{table}[h]
\centering
\small
\setlength{\tabcolsep}{4pt}
\caption{CAS by register with 95\% bootstrap intervals over rules and the number of rules in each
stratum. Rule counts are the frozen probe set, 295 + 26 + 46 + 11 + 19 = 397; a cell's CAS
denominator is at or below its column count, because rules whose probes all returned
\textsc{na} are excluded per target. Registers are never pooled: 295 vendor-specification
rules would otherwise dominate the aggregate. The intervals are what the strata support: GPT-4o and Gemini~2.5~Flash separate vendor
specification from the regulatory registers with non-overlapping intervals, while the Claude
models' spread across registers does not clear its own uncertainty.}
\label{tab:byclass-app}
\begin{tabular}{@{}lccccc@{}}
\toprule
\textbf{Target} & \textbf{Vendor spec} & \textbf{Analytical} & \textbf{Conduct reg.} & \textbf{AUP} & \textbf{Statute} \\
& $n{=}295$ & $n{=}26$ & $n{=}46$ & $n{=}11$ & $n{=}19$ \\
\midrule
Haiku 4.5 & 0.849 & 0.917 & 0.792 & 0.728 & 0.748 \\
 & \tiny[0.81,0.88] & \tiny[0.83,0.99] & \tiny[0.71,0.87] & \tiny[0.52,0.90] & \tiny[0.63,0.86] \\
Sonnet 4.5 & 0.835 & 0.928 & 0.795 & 0.715 & 0.780 \\
 & \tiny[0.80,0.87] & \tiny[0.82,1.00] & \tiny[0.71,0.87] & \tiny[0.54,0.87] & \tiny[0.64,0.90] \\
GPT-4o & 0.708 & 0.587 & 0.494 & 0.473 & 0.461 \\
 & \tiny[0.66,0.75] & \tiny[0.47,0.72] & \tiny[0.41,0.58] & \tiny[0.23,0.71] & \tiny[0.34,0.58] \\
Gemini 2.5 Flash & 0.657 & 0.565 & 0.514 & 0.517 & 0.467 \\
 & \tiny[0.62,0.69] & \tiny[0.45,0.68] & \tiny[0.42,0.60] & \tiny[0.36,0.69] & \tiny[0.34,0.59] \\
\bottomrule
\end{tabular}
\end{table}

\paragraph{Corpus sensitivity, and the headline on the conservative set.}
Every reported result runs on the 397 rules of the frozen probe set, drawn from the 401
majority-bindable obligations that survive the scope exclusion. Exactly four screened rules
never entered the probe set and no probed rule falls outside the screened set; the four are
listed with their exclusion reason in the released artifact. The stricter stages of the
funnel are not merely reported as properties of the criterion. Of the 397 probed rules, 172 also
belong to the 231-rule non-mixed conservative set, and every quantity in
Section~\ref{sec:results} was recomputed on that subset.

Nothing that the paper concludes depends on the looser threshold. Target ordering on both
specifications is identical across the two admission sets, and every interval on the stricter
set contains the corresponding majority-set point estimate. Absolute CAS moves by between
$-0.025$ and $+0.051$ and not in a single direction: it rises for the weaker arms on both
documents and falls slightly for the Claude models on the Constitution, which is what a
threshold that removes contested rules should do to a cell that had few of them to lose.

The own-specification advantage does not disappear under the stricter criterion and does not
become resolved either. Against the three-arm baseline it moves from $+3.8$, $+2.5$ and $-2.7$
points for Haiku, Sonnet and GPT-4o on the majority set to $+4.0$, $+5.0$ and $-7.4$ on the
conservative set, with the sign stable for all three targets on both. \emph{We report the sign
as stable across admission thresholds and the magnitude as resolved under neither}, because the
conservative Constitution arm carries 31 rules and the baseline the difference is taken against
is estimated from three arms at that same scale. What the stricter threshold does not do is
manufacture an effect the looser one hid, which is the objection it was raised against.

Table~\ref{tab:xvendor} gives CAS by target and specification,
Table~\ref{tab:byclass-app} by register, and Figure~\ref{fig:radar} plots the suite profile for
the three targets whose profile is complete on every axis the figure plots.

\paragraph{Coverage.}
Four arms cover the full corpus: Haiku 4.5, Sonnet 4.5, GPT-4o and Gemini~2.5~Flash, at 1{,}588
probe-response pairs each. The headline results rest on these four and on the paired GPT-5.5
comparison; the difficulty baseline additionally uses the two arms below. Gemma 2 9B and Llama 3.1 8B each completed both published
specifications at full coverage, which is what qualifies them as difficulty-baseline arms. As Appendix~\ref{app:swap} records, Gemma~2~9B carries the widest judge-attribution
band of any arm at 0.216 and Llama~3.1~8B the third widest at 0.162, so neither supports a claim that turns on an absolute
level. A Gemini~2.5~Pro arm was attempted and is excluded in full, because it never ran. The
provider returned HTTP 429 quota-exhaustion errors on 544 of the 546 requests that reached it,
across two launches on separate days, each abandoned by the runner after eight consecutive
failures with zero successes. No response from that target exists to score, so the exclusion
raises no selection question: it is the absence of data rather than a choice among data. The
stored error rows are retained in the artifact.

\end{document}